%% file: query-intent.tex
\documentclass[sigconf]{acmart}

\def\BibTeX{{\rm B\kern-.05em{\sc i\kern-.025em b}\kern-.08emT\kern-.1667em\lower.7ex\hbox{E}\kern-.125emX}}

\copyrightyear{2020}
\acmYear{2020}
\setcopyright{acmlicensed}
\acmConference[CIKM]{The Conference on Information and Knowledge Management}{2020}{Galway, Ireland}
\setcopyright{none}
\renewcommand\footnotetextcopyrightpermission[1]{}
\usepackage{pifont}
\usepackage{multirow}

\usepackage{subcaption}
\usepackage{courier}
\usepackage{amsthm}

\theoremstyle{plain}

\theoremstyle{definition}

\theoremstyle{remark}

\usepackage{amssymb}

\begin{document}

\title{Deep Search Query Intent Understanding}

\author{Xiaowei Liu, Weiwei Guo, Huiji Gao, Bo Long}
\affiliation{
  \institution{LinkedIn, Mountain View, California}
}
\email{{xwli, wguo, hgao, blong}@linkedin.com}

\begin{abstract}
    Understanding a user's query intent behind a search is critical for modern search engine success. 
    Accurate query intent prediction allows the search engine to better serve the user's need by rendering results from more relevant categories. 
    This paper aims to provide a comprehensive learning framework for modeling query intent under different stages of a search. We focus on the design for 1) predicting users' intents as they type in queries on-the-fly in typeahead search using character-level models; and 2) accurate word-level intent prediction models for complete queries. Various deep learning components for query text understanding are experimented. Offline evaluation and online A/B test experiments show that the proposed methods are effective in understanding query intent and efficient to scale for online search systems.
    

\end{abstract}
\keywords{Query Intent, Query Classification, Natural Language Processing, Deep Learning}

\maketitle

\input{intro.tex}

\input{qi-linkedin.tex}

\input{problem}
\input{model.tex}

\input{experiments.tex}
\input{lesson}

\input{related-work}

\vspace{-1mm}
\section{Conclusion}
This paper proposes a comprehensive framework for modeling the query intent in search systems for different product components. The proposed deep learning based models are proven to be effective and efficient for online search applications. 
Discussions about the challenges for deploying these models to  production as well as our insights in making these decisions are provided. We hope the framework as well as the experiences during our journey could be useful for readers designing real-world query understanding and text classification tasks.

\bibliographystyle{ACM-Reference-Format}
\bibliography{query-intent}

\end{document}

%% file: intro.tex
\section{Introduction}
Modern search engines provide search services specialized across various domains (\textit{e.g.}, news, books, and travel).
Users come to a search engine to look for information with different possible intents: choosing favorite restaurants, checking opening hours, or restaurant addresses on Yelp; searching for people, finding job opportunities, looking for company information on LinkedIn, etc.    
Understanding the intent of a searcher is crucial to the success of search systems. Queries contain rich textual information provided explicitly by the searcher, hence a strong indicator to the searcher’s intent. 
Understanding the underlying searcher intent from a query, is referred to the task of query intent modeling.

\begin{figure*}
\includegraphics[width=\linewidth]{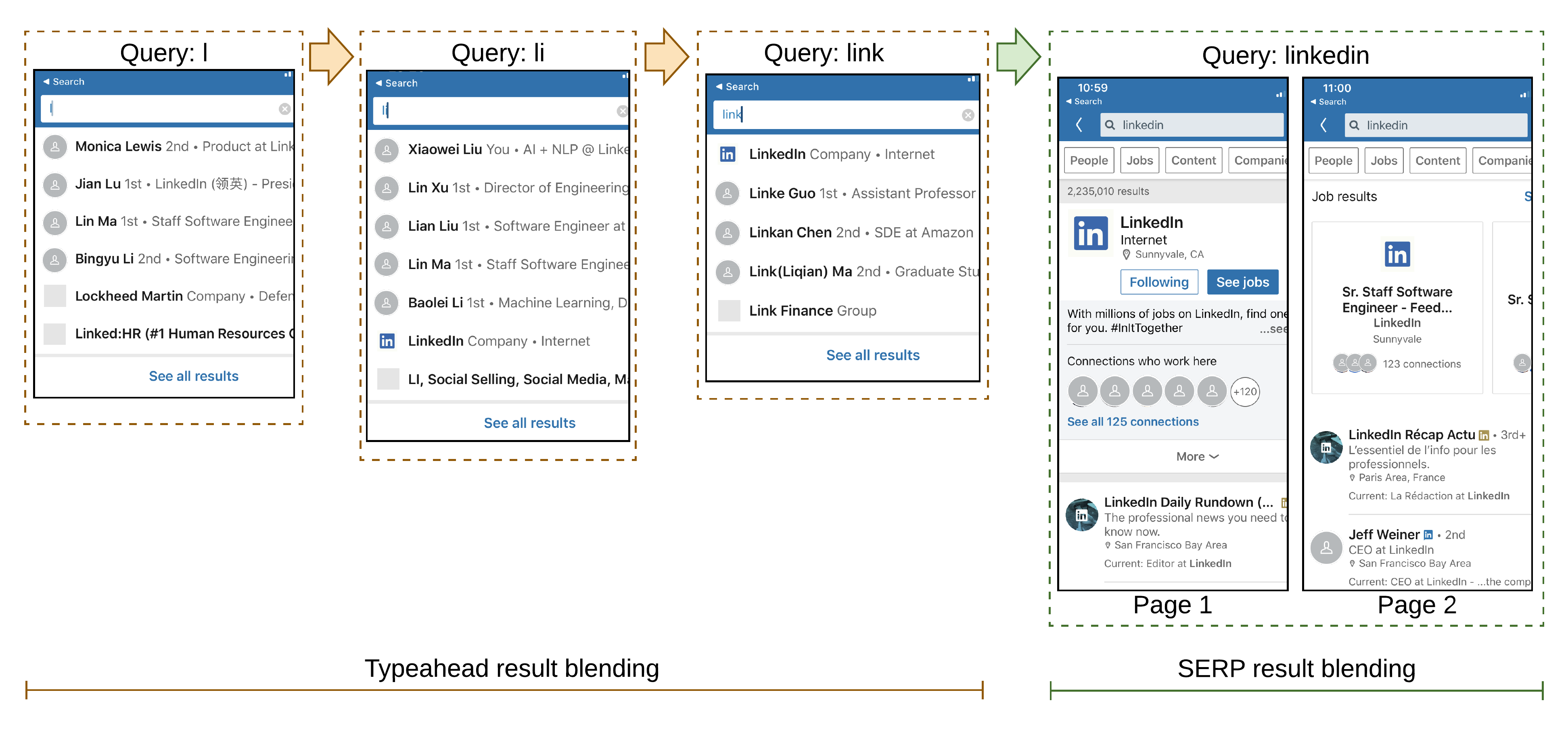}
\caption{Query intent in search engines for incomplete queries (typeahead blending) and complete queries (SERP blending).} 
\label{figure:blending-online}
\end{figure*}

Query intent is an important component in the search engine ecosystem \cite{kang2003query,li2008learning,hu2009understanding}. As shown in Figure \ref{figure:blending-online}, when the user starts typing a query, the intent is predicted based on the incomplete character sequence; when the user finishes typing the whole query, a more accurate intent is predicted based on the completed query.  Understanding the user intent accurately allows the search engine to trigger corresponding vertical searches, as well as to better rank the retrieved documents based on the intent \cite{arguello2009sources}, so that users do not have to refine their searches by explicitly navigating through the different facets in the search engine. 

Traditional methods rely on bag-of-words representation and rule based features to perform intent classification \cite{arguello2009sources, ashkan2009classifying}. Recently, deep learning based models \cite{hashemi2016query} show significant improvement, which can handle similar words/word sense disambiguation well. 
However, developing deep learning based query intent models for productions requires considering several challenges.  Firstly, production models have very strict latency requirements, and the whole process needs to be finished within tens of milliseconds.
Secondly, queries, usually with two or three words in a complete query or several characters in a incomplete one, have limited contexts. 

This paper proposes a practical deep learning framework to tackle the two challenges, with the goal of improving LinkedIn's commercial search engine. Two search result blending components were identified where query intent is useful: incomplete query intent for typeahead blending, and complete query intent for SERP blending (search engine result page). 

The common part of both systems is to use query intent to assist the ranking of retrieved documents of different types. Meanwhile, the two products have their unique challenges. Typeahead blending has strong latency requirements; the input is an incomplete query (a sequence of characters); and it is okay to return a fuzzy prediction, since users will continue to type the whole query if he/she does not find the results relevant. On the other hand, SERP blending has less latency constraint compared to typeahead but a higher accuracy requirement as it directly affects the search result page. 

Based on the characteristics of production applications, we propose different solutions. For typeahead blending, character-level query representation is used as the resulting models are compact in terms of the number of parameters. Meanwhile, it can handle multilinguality well due to the small vocabulary size. For SERP blending, the complete query intent model is word level. Since accuracy is a high standard, BERT is explored to extract query representations which lead to a more robust model.

This paper is motivated by tackling the challenges in query intent prediction, while satisfying production needs in order to achieve real-world impact in search systems. The major contributions are:

\vspace{-3mm}

\begin{itemize}
    \item Developed a practical deep query intent understanding framework that can adapt to various product scenarios in industry. It allows for fast and compact models suitable for online systems, with the ability of incorporating traditional features that enables more accurate predictions.
    \item Developed and deployed character-level deep models to production that is scalable for incomplete query intent prediction. In addition, we propose a multilingual that incorporates language features which is accurate and easier to maintain than traditional per-language models.
    \item Developed and deployed a BERT based model to production for complete query intent prediction. To our best knowledge, this is the first reported BERT model for query intent understanding in real-word search engines.
    \item Conducted comprehensive offline experiments and online A/B tests on various neural network structures for deep query intent understanding, as well as in-depth investigation on token granularity (word-level, character-level), DNN components (CNN, LSTM, BERT), and multilingual models, with practical lessons summarized.
\end{itemize}

%% file: qi-linkedin.tex
\section{Query Intent Understanding at LinkedIn}
LinkedIn search hosts many different types of documents, 
\textit{e.g.}, user profiles, job posts, user feeds, etc. When a user issues a query without specifying the document type they are interested in, identifying the intent is crucial to retrieve relevant documents and provide high-quality user experience.  At LinkedIn, we define the query intent as the document type.

Query intent is important for result blending \cite{hu2009understanding}: (1) When an intent is not presented in the query, the corresponding vertical search may not be triggered. (2) For the documents retrieved from the triggered vertical searches, a result blending algorithm will rank the documents based on detected intents and other features.  In this section, we present two productions where query intent is an important feature for the blending algorithm, followed by how query intent is used in the blending algorithm.

\subsection{Query Intent in Typeahead Blending}
When a user starts typing, the query intent is detected and used in typeahead blending. At LinkedIn, the typeahead product directly displays document snapshots from multiple vertical searches, which is different from traditional query auto completion that only generates query strings. 
The left three snapshots in Figure \ref{figure:blending-online} shows an example of typeahead blending results at LinkedIn.  The example assumes the user is searching for the company "LinkedIn". Blended results are rendered as soon as the user typed one letter "l". Next, given the query prefix "li", the intent prediction has a tendency towards a people result type and many people profiles are ranked higher than company or groups results. After the user types "link" in the third picture, the company result LinkedIn is ranked first.

Query intent for typeahead blending is challenging given that the queries are often incomplete and contains only several letters. In addition, for every keystroke, the system needs to retrieve the documents from different vertical searches and blend the results.  It means the query intent models will be called frequently, and each run should be finished within a short time.

\vspace{-1mm}
\subsection{Query Intent in SERP Blending}
SERP blending is a more common component in search engines than typeahead blending.
When a user finishes typing and hits "search", a complete query is issued; the query intent is identified and used for retrieved document blending.
The right most block in Figure \ref{figure:blending-online} shows SERP blending results for a complete query "linkedin", including company pages, people profiles, job posts, etc.  

Compared to typeahead blending, query intent in SERP blending has a larger latency buffer. Queries contain complete words, however, it still suffers from limited contexts: only several words in a query for intent prediction.

\subsection{Retrieved Document Blending Systems}
Both typeahead blending and SERP blending systems follow a similar design.  Multiple features are generated for  blending/ranking the retrieved documents: (1) Probability over each intent that is based on query texts and user behaviors (the query intent model output); (2) matching features between the query and retrieved documents; (3) personalized and contextualized features.

In the rest of this paper, we focus on how to generate the query intent probability for typeahead blending and SERP blending.

%% file: problem.tex
\section{Problem Definition}
Both incomplete query intent and complete query intent are essentially classification tasks. Without loss of generosity, given a user id $u \in \mathbb{U}$ and a query string $q \in \mathbb{Q}$, the goal is to learn a function $\gamma$ predict the predefined intent $i$ in the finite number of intent classes $\mathbb{I} = \{i_1, i_2, ... i_n\}$:
\begin{align*}
    \gamma:  \langle\mathbb{Q}, \mathbb{U}\rangle \rightarrow \mathbb{I}
\end{align*}

For the two tasks, incomplete/complete query intent, the intent class sets are slightly different, due to the design of products. As shown in the next section, deep learning models are applied to query strings; the user ids are used to generate personalized features.

%% file: model.tex
\section{Deep Query Intent Understanding}
 
In this section, we introduce the proposed intent modeling framework, as well as the detailed design of two applications: incomplete query intent model and complete query intent model.
\subsection{Product Requirements}
As shown in Figure \ref{figure:blending-online}, there are two result blending products that rely on query intent: typeahead blending and SERP blending.  These two products pose several requirements of the query intent models.

Typeahead blending has a strict latency standard: for every keystroke, the model needs to return the results. Meanwhile, it does not require very high accuracy since the prediction becomes more precise as it receives more characters. On the other hand, SERP blending has more latency buffer, and it requires high accuracy of query intent, otherwise users might abandon the search.




\subsection{Intent Modeling Framework}

\begin{figure}
  \includegraphics[width=0.5\textwidth]{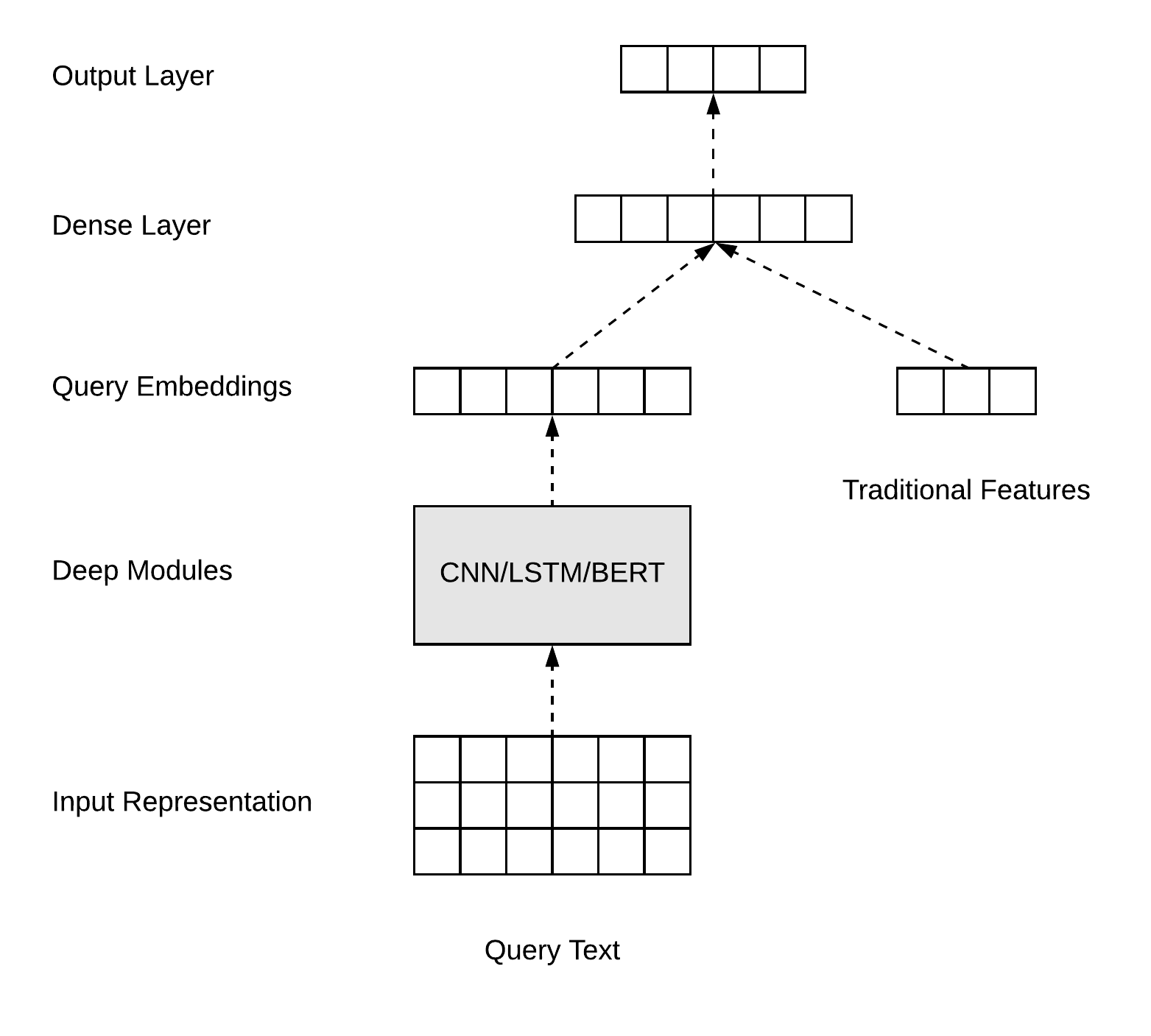}
  \vspace{-4mm}
  \caption{Deep query intent understanding framework.}
  \label{figure:framework}
\vspace{-5mm}
\end{figure}

Driven by the product requirements, we design a framework for query intent understanding. The overall architecture is in Figure \ref{figure:framework}.

\subsubsection{Input Representation}

The input to the model is represented in a sequence of embeddings. Two granularity choices are provided: character- and word-level embeddings to support incomplete and complete queries, respectively.


\subsubsection{Deep Modules}

In this framework, several popular text encoding methods are provided to generate query embeddings. This enables good flexibility to adapt the framework to various product scenarios under different latency / accuracy constraints.

\vspace{1mm}
\textbf{CNN} is powerful at extracting local ngram features in a sequence \cite{kalchbrenner2014convolutional,kim2014convolutional}. The input to the CNN is a sequence of token embeddings, i.e. the embedding matrix. The 1-dimensional convolution layers could involve multiple filters of different heights. The width of the CNN filters is always the same size as the embedding dimension, while the height of the filters could vary—it represents word or character n-grams covered by the filters. Max-pooling over time is done after each convolution layer. 

\vspace{1mm}
Compared with CNN, the long distance dependencies can be better captured by \textbf{LSTM} \cite{hochreiter1997long}, especially the character sequence. Bi-directional LSTM \cite{zhou2016text} is used to model the sequence information from both forward and backward. The last hidden state of both layers are concatenated together to form the output layer.

\vspace{1mm}
\textbf{BERT} \cite{devlin2018bert} uses self attention \cite{vaswani2017attention} to explicitly integrate contextual word meaning into the target word, hence it is better at word sense disambiguation.  Meanwhile, the pretraining enables using a large amount of unsupervised data.  Given a query, BERT takes a sequence of tokens as input and output the contextualized representation of the sequence. A special token \texttt{[CLS]} at the beginning of each sequence models the representation of the entire sequence and is used for classification tasks. 

\subsubsection{Traditional Features}
Traditional features are hand-crafted features, which are powerful for capturing contextual information that is complementary to the deep textual features. There are various types of traditional features that can be considered in production, such as language features, user profile / behavioral features. These features are especially important for enhancing the limited context for short queries.
In a wide-and-deep fashion~\cite{cheng2016wide}, traditional features are concatenated with the query embeddings, and then fed to a dense layer to get non-linear interactions among the features. 
\subsection{Incomplete Query Intent Modeling}

In typeahead search, users usually type a query prefix, and select the results from the drop-down bar. In this case, a large number of incomplete words are generated, which are recognized as out-of-vocabulary words.  This motivates us to design the incomplete query intent classification with character-level representations.  The character-level models have additional benefits: (1) it is more \textbf{robust} to spelling errors, compared to word-level models where words with wrong spelling will be out of vocabulary; (2) the resulting model is \textbf{compact} (1.4 Megabytes with 500 characters), as the character vocabulary size is small. 

Due to the ability of capturing long range dependency information, LSTM is best suited for this problem, as the sequence could be over 10 characters.  In contrast, CNN captures the n-gram letter patterns (e.g., tri-letters). However, it does not keep track of what are the contexts of the tri-letters, as well as the position of the tri-letters. 
BERT is not applied in this task, because the typeahead product has a strict latency constraint. The latency would not meet the production constraint for BERT is too computationally heavy.

\begin{figure*}

\begin{subfigure}{0.38\textwidth}
\hspace{1cm}
\includegraphics[width=\linewidth]{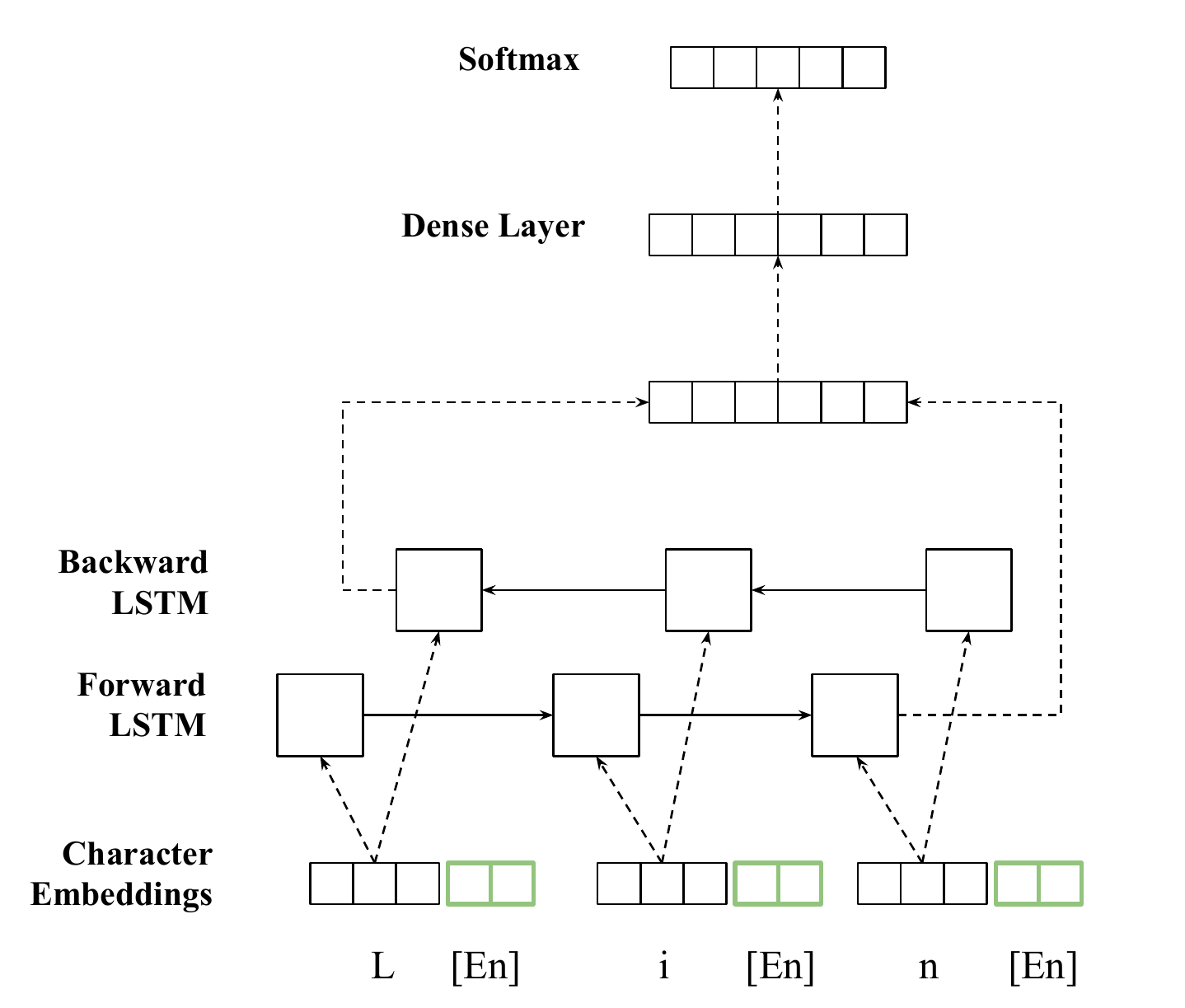}
\hspace*{\fill} 
\caption{Embedding.} \label{figure:iqi-embed}
\end{subfigure}
\begin{subfigure}{0.4\textwidth}
\hspace{2cm}
\includegraphics[width=\linewidth]{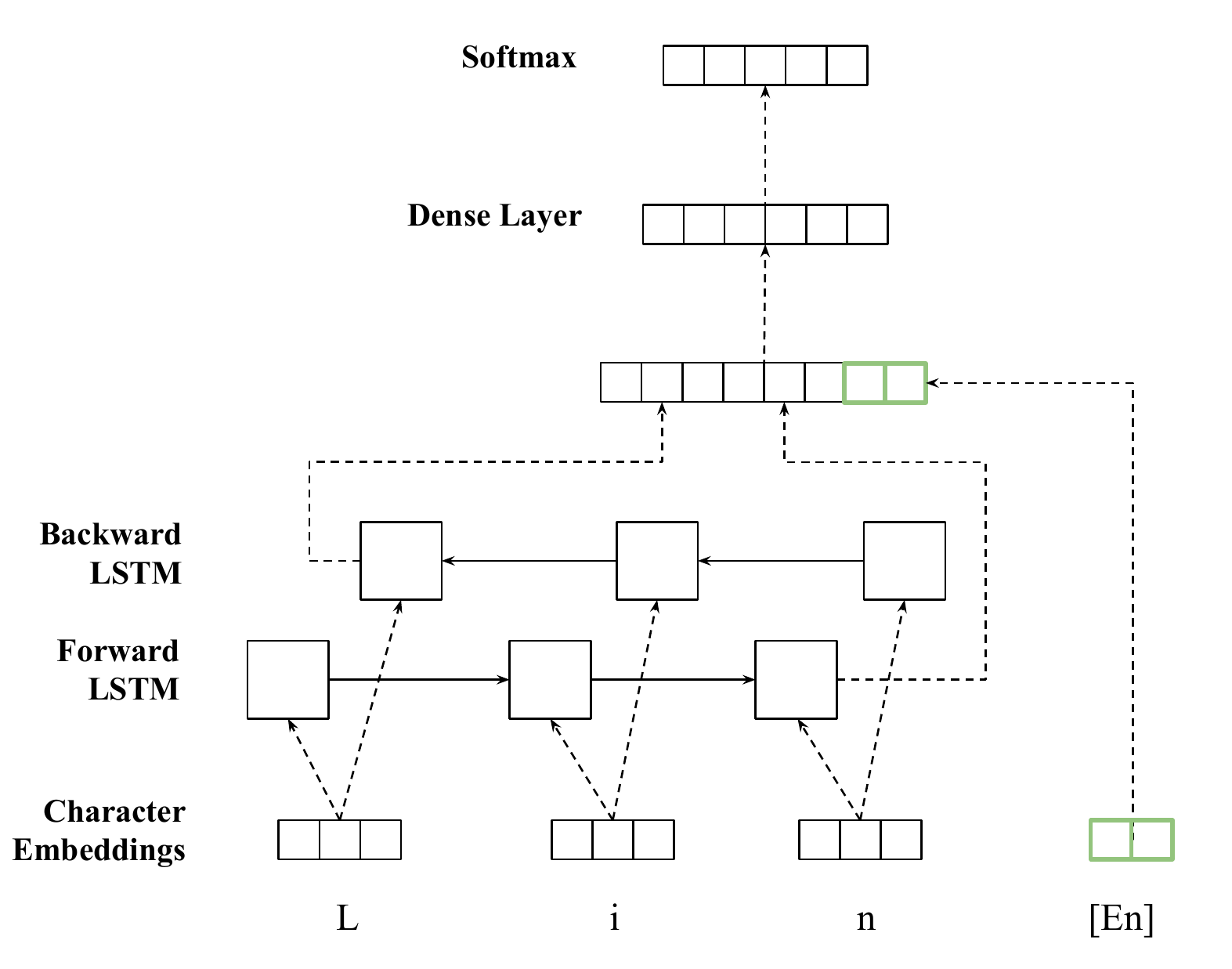}
\vspace{1mm}
\caption{Concatenation.} \label{figure:iqi-concat}
\end{subfigure}
\hspace*{\fill} 

\caption{The language feature embedding and concatenation scheme for multilingual character-level query intent model.} 
\label{figure:iqi-multilingual}
\vspace{-1mm}
\end{figure*}

\subsubsection{Multilingual Support}
\label{section:multilingual}
\noindent Search systems often experience a large amount of traffic in many countries with different languages. In LinkedIn typeahead search, more than 30\% of search traffic is from languages other than English. Handling the performance gap between English and other languages is crucial to enable a good user experience in typeahead.

The typical way to handle multilinguality is to train per-language models, which is not easy to maintain and scale. The character-level approach can be easily extended to support multiple languages with a unified model, since the character vocabulary of all languages is much smaller than the word vocabulary.

To train a unified multilingual model, two approaches are designed to model the language/locale information:


\noindent\textbf{Language Feature Embedding: }
The first approach is to concatenate the interface locale embedding to the input query embedding, and then feed it through a BiLSTM to generate interface locale “embedded” query features, as shown in Figure \ref{figure:iqi-embed}. The embeddings of the interface locales are randomly initialized and together trained with the network.

\noindent\textbf{Language Feature Concatenation: }
The second approach is to use the BiLSTM layer for extracting query features from query embeddings, and then concatenate the interface locale embeddings to the last output state of the BiLSTM, as in Figure \ref{figure:iqi-concat}. Compared to the first approach, BiLSTM is unaware of the language.


\subsection{Complete Query Intent Modeling}

We design word-level CNN, LSTM, and BERT based models for complete query intent classification. 
CNN is good at capturing word n-gram patterns. LSTM is powerful for modeling long sequence context. BERT further enriches the query representation by modeling the context meaning. Specifically, a pre-trained BERT model with LinkedIn data (\textbf{LiBERT}) is used.

There are two major motivations on pre-training a BERT model with in-domain LinkedIn data (\textbf{LiBERT}) instead of using the off-the-shelf BERT models released by Google \cite{devlin2018bert}:

\vspace{1mm}
\noindent\textbf{Latency Constraint.} 
The probabilities of different intents given a query need to be computed on-the-fly instead of pre-computing as there are unique queries from different users searched each day. Given that GPU/TPU serving is not yet available at LinkedIn, the latency for computing the query intent outputs using the BERT-Base model from Google won’t satisfy the critical latency requirement. The LiBERT model has a lighter architecture (discussed in Section \ref{section:libert-pretrain}) than the smallest model released from the original paper~\cite{devlin2018bert}.

\vspace{1mm}
\noindent\textbf{Better Relevance Performance.} In addition to the benefit of low latency, pre-training the BERT model from scratch with LinkedIn data leverages the in-domain knowledge in the corpus, therefore can provide better relevance performance for downstream fine-tuning tasks. An example is that \texttt{LinkedIn}, \texttt{Pinterest} are both out-of-vocabulary words in Google’s BERT-Base model. These company names are generally important for LinkedIn search.


Note that we haven't explored a multilingual approach for the word-level complete query intent models. 
This is because the word-level vocabulary could be fairly large if the major languages are included and therefore adding more complexity to the models.

%% file: experiments.tex
\section{Experiments}
Experiments on both incomplete and complete query intent prediction models are discussed in this section. Performance comparison in both offline and online metrics are shown by relative differences instead of absolute values due to company policy. We also share our analysis in the scalability of these deep learning based models, as well as the online A/B testing performance of these models compared with production baselines. 
Tensorflow serving with CPU is used for online inference for the deep models. Traditional features such as the user behavioral features are pushed to a online store after daily offline computation and can be accessed during model inferencing. 
All the reported online metrics lift are statistically significant ($p<0.05$) and are collected based on 4 weeks of 50\% A/B testing search traffic. The LSTM model and BERT model are fully ramped to production after the A/B testing for incomplete query intent and complete query intent, respectively.

\subsection{Training Data Collection}
\label{section:label-data-collection}
Search click-through log is used for training data collection. For example, given that a user searched for \textit{linkedin sales solutions} and clicked on a job posting instead of a \textit{people} or \textit{content} page, a \textit{job} intent label is assigned to this query. Similarly, when a user typed an incomplete query \textit{face} and chose the \textit{facebook} company page, then a \textit{company} intent is inferred.


For both incomplete and complete intent models, we collected training data sampled from clicked-through log for a month. For train/dev/test dataset, we sampled 42M/21K/21K for English incomplete models, 63M/36K/36K for multilingual incomplete query models, and 24M/50k/50k for complete query intent models.

\subsection{Incomplete Query Intent Prediction}

\subsubsection{Experiment Setting}


We evaluate the performance for character-based CNN and LSTM models in the offline experiments. 
The baseline model is learnt using letter tri-gram bag-of-words features with a logistic regression classifier. 

In the CNN/LSTM based models, character embeddings of size 128, with a 500 vocabulary size are used. 
The characters in the vocabulary are extracted from the most frequent characters in the training data. The embeddings are randomly initialized at the beginning of training. We used 128 convolutional filters of size 3 for training the CNN model. 
In the Bidirectional LSTM model, the same character vocabulary is used as in the CNN model and the number of hidden units is 128.

\subsubsection{Offline Experiments}
Both English and multilingual incomplete query models are investigated in offline experiments. Throughout experiment result comparison, we use the $-en$ and $-i18n$ notation to represent English models and multilingual models.
 
\begin{table}
\small
\caption{Offline performance comparison for different character-level models vs production baseline.}
\label{table:iqi-offline}
\begin{tabular}{lccc}
\toprule
\textbf{Model}     & \textbf{Accuracy}    & \textbf{F1 (people)}     & \textbf{F1 (company)}    \\
\midrule
Tri-letter & - & -  & -  \\
CNN-en               & $+4.14\%$ & $+2.69\%$ & $+11.39\%$ \\
LSTM-en            & $+7.11\%$  & $+4.46\%$ & $+20.36\%$  \\
\bottomrule
\end{tabular}
\vspace{-1mm}
\end{table}

\vspace{1mm}
\noindent\textbf{English Model.}
The first set of experiments were conducted on English incomplete queries. Experimental results show that compared with the tri-letter model, the CNN-based model CNN-en is able to achieve a higher accuracy, which is mostly because of the fact that CNN could extract more abstract word pattern features than traditional bag-of-words features. As shown in Table~\ref{table:iqi-offline}, the Bidirectional LSTM model LSTM-en can further improve the performance (+7.11\% vs. +4.14\%), since LSTM can better model the long range character sequences. 

\begin{table}
\small
\caption{The size of the character vocabulary for incomplete query intent models.}
\label{table:iqi-vocab}
\begin{tabular}{lcccc}
\toprule
\textbf{Model}   &\textbf{Vocab}    & \textbf{Accuracy}    & \textbf{F1 (\textit{people})}     & \textbf{F1 (\textit{company})}    \\
\midrule
Tri-letter          & - & - &-  & -\\
CNN-en                  & $70$ & $+4.09\%$  &  $+2.69\%$ & $+11.38\%$ \\
CNN-en                  & $500$ &  $+4.14\%$ & $+2.73\%$  & $+11.75\%$\\
LSTM-en                 & $70$ &  $+7.04\%$ & $+4.46\%$  & $+20.36\%$\\
LSTM-en                 & $500$ &  $+7.11\%$ & $+4.49\%$ & $+20.49\%$\\
\bottomrule
\end{tabular}
\vspace{-1mm}
\end{table}

We also compare the impact of character vocabulary size for the deep  models (Table \ref{table:iqi-vocab}). It is interesting to see that even with a small vocabulary (only 70 characters consisting of lower-case English characters and special characters), the accuracy is significantly higher compared with the baseline model.
\vspace{1mm}

\noindent\textbf{Multilingual Model.} In the multilingual experiments, we first collect training data using a similar procedure on top 30 international locales, such as French, Portuguese, German, etc. The international locales are used as language feature in our multilingual model. The most frequent 10k characters are extracted as vocabulary.

\begin{table}
\small
\caption{Multilingual models for character-level incomplete query intent.}
\label{table:iqi-multi}
\begin{tabular}{lcccc}
\toprule
\textbf{Model}       & \textbf{Accuracy}    & \textbf{F1 (\textit{people})}     & \textbf{F1 (\textit{company})}    \\
\midrule
LSTM-i18n                  & -  & -  & - \\
LSTM-i18n-embed            & $+0.67\%$ & $+0.44\%$  & $+1.34\%$  \\
LSTM-i18n-concat           & $+0.55\%$ & $+0.37\%$  & $+1.12\%$  \\
\bottomrule
\end{tabular}
\vspace{-1mm}
\end{table}

Table~\ref{table:iqi-multi} summarizes the performance of two different strategies discussed in Section~\ref{section:multilingual}. 
The results suggest that both methods outperform the baseline  without language embedding injected. The language feature embedding method (LSTM-i18n-embed) outperforms the concatenation (LSTM-i18n-concat) method slightly. 
This might be that in LSTM-i18n-concat, the LSTM is aware of the language, hence generating a more meaningful representation for the character input sequence. 

We further compare the effectiveness of the multilingual models to the traditional per-language models. Table\ref{table:iqi-multi-single} shows the overall accuracy and per-language accuracy. The LSTM-per-lang is trained and evaluate on each individual language portion. LSTM-i18n is the multilingual model without language features added.  LSTM-i18n-embed is with language feature added to the character embeddings. The overall accuracy for LSTM-i18n-embed is comparable to the per-language model. We also sampled 4 languages to show the individual performances. For French (fr) and Chinese (zh), both multilingual models improved performance compared with the single language LSTM-i18n-per-lang model trained on data from single language only. This is true that for certain languages, multilingual helps the model to learn from other languages and generalize better to help predictions on the specific languages. From results in the other 2 languages, both LSTM-i18n and LSTM-i18n-embed did not outperform the per-language models. This is often expected. But with the language feature embedded (LSTM-i18n-embed), we observed an increased accuracy compared to the LSTM-i18n model.

\begin{table}
\small
\caption{Multilingual model vs per-language models. } 
\label{table:iqi-multi-single}
\begin{tabular}{lccccc}
\toprule
Model & Overall &	fr &	de &	zh &   	da  \\
\midrule
LSTM-i18n-per-lang &	- &	- &	- &	-  & -\\
LSTM-i18n & $-0.65\%$ &  $+0.28\%$   & $-0.84\%$   & $+0.54\%$ & $-1.24\%$	 \\
LSTM-i18n-embed & $+0.02\%$ &	 $+0.56\%$   & $+1.01\%$   & $-1.11\%$ & $-0.66\%$	 \\
\bottomrule
\end{tabular}
\end{table}

\subsubsection{Online Experiments}


\begin{table}
\small
\caption{Online performance comparison for incomplete query intent models vs baseline linear model.}
\label{table:iqi-online}
\begin{tabular}{lccc}
\toprule
\textbf{Model} & \textbf{Traffic} & \textbf{Metrics} & \textbf{Lift} \\
 \midrule
\multirow{2}{*}{LSTM-en} & \multirow{2}{*}{English} & Search session success & +0.43\% \\
 &  & Time to success (mobile) & -0.15\% \\
\multirow{2}{*}{LSTM-i18n-embed} & \multirow{2}{*}{Non-English} & Search session success & +0.86\% \\
 &  & Time to success (mobile) & -0.19\% \\
 \bottomrule
\end{tabular}
\end{table}

As shown in Table~\ref{table:iqi-online}, the English model (LSTM-en) and multilingual model (LSTM-i18n-embed) are tested in typeahead search production with English/non-English traffic, respectively. 
The online business metrics measure the success criteria within search sessions. 
Both models show significant performance gains for \textit{search session success}. An interesting observation is that the \textit{average time to success} on mobile is reduced for both models. An intuitive reason is that mobile users are more likely to be click on the high quality results rendered in typeahead search due to the typing behavior compared with on desktop. 

\subsection{Complete Query Intent Prediction}
\label{section:complete-intent}
\subsubsection{Experiment Setting}

The input is word-level embeddings. The embeddings are initialized using GloVe\cite{pennington2014glove} word vector representations pre-trained on LinkedIn text data. It covers a vocabulary of 100K with dimension 64. 

The production baseline model is a logistic regression model, with bag-of-words features, and user profile/behavioral features, etc. In the offline experiments, we first compare the performance among the deep learning models with the production baseline. 
For the CNN-based model, we used 128 filters of height 3 (tri-gramd) for the 1-D convolution. The hidden state size is 128 for the LSTM-based model. For both CNN and LSTMs, after the query embedding is generated, a layer of size 200 is used to capture the non-linear interactions between query representation and traditional features.

\subsubsection{LiBERT Pre-training}
\label{section:libert-pretrain}

\begin{table}
\small
\caption{Model architecture comparison between LiBERT Google's BERT-Base.}
\label{table:libert-param}
\begin{tabular}{lcc}
\toprule
\textbf{BERT HParams} & \textbf{BERT-Base} & \textbf{LiBERT} \\
\midrule
Layers & 12 & 3 \\
Hidden size & 768 & 256 \\
Attention heads & 12 & 8 \\
\midrule
Total \#params &  110M & 10M  \\
\bottomrule
\end{tabular}
\end{table}

The LiBERT-baesed model is fine-tuned with a BERT model pre-trained on LinkedIn data.
The pre-training data include a wide variety of data sources across LinkedIn: search queries, user profiles, job posts, etc. The collected data for pre-training include around 1 billion words. 
A light-weight structure is used compared to the BERT-Base (a smaller model published in BERT \cite{devlin2018bert}). A comparison between BERT-Base and LiBERT is given in Table \ref{table:libert-param}. 

\subsubsection{Offline Experiments}

\begin{table}
\small
\caption{Offline performance comparison for the word-level models vs production baseline.}
\label{table:qim-offline}
\begin{tabular}{lccc}
\toprule
\textbf{Model}     & \textbf{Accuracy}    & \textbf{F1 (\textit{people})}     & \textbf{F1 (\textit{job})}    \\
\midrule
LR-BOW & - & -  & - \\
\midrule
CNN-word                & $+6.40\%$ & $+1.36\%$ & $+17.35\%$ \\
CNN-char        & $+3.49\%$ & $-0.50\%$ & $+10.29\%$\\
LSTM-word              & $+6.69\%$  & $+1.47\%$ & $+17.99\%$  \\
LSTM-char           & $+4.63\%$ & $+0.23\%$ & $+13.46\%$  \\
\midrule
BERT-Base              & $+8.20\%$  & $+2.33\%$ & $+19.33\%$  \\
LiBERT              & $+8.35\%$  & $+2.46\%$ & $+20.60\%$  \\
\bottomrule
\end{tabular}
\end{table}

Table \ref{table:qim-offline} summarizes the performance of different models.  
CNN/LSTM outperforms the baseline models, by $+6.40\%$/$+6.69\%$, respectively. 

We further compared the performances of the complete query models in different token granularity and deep model types.
First, the word-level models outperforms character-level models consistently. This implies that for complete queries, word level representations captures more meaningful features than character sequences. Next, we compare the performance in LSTM and CNN models under different token granularity. More specifically, on word-level models, LSTM and CNN have similar performances. This implies that the useful sequential information is limited in the word sequences due to the short length of the queries and possibly the effectiveness of word-level information such as word meaning. This is further proved as when comparing the character-level models, LSTM outperforms CNN by a larger margin, meaning that longer range information is more useful in character sequence modeling. Similar observation can be found in Table~\ref{table:iqi-offline} on incomplete query intent model experiments, where we saw even more significant improvement on LSTM over CNN, implying that the LSTM is more suitable in modeling the incomplete character sequences in typeahead than the complete queries in SERP.



BERT models (BERT-Base and LiBERT) can further improve the accuracy over CNN/LSTM  models. This can be attributed to the contextual embeddings that are able to disambiguate words under different contexts better.

\begin{table}
\small
\caption{Performance comparison of whether to incorporate traditional features.}
\label{table:qim-trad-feat}
\begin{tabular}{lcccc}
\toprule
\textbf{Model}  & \textbf{Trad Features}   & \textbf{Accuracy}    & \textbf{F1 (\textit{people})}     & \textbf{F1 (\textit{job})}    \\
\midrule
LR-BOW & - & - & -  & - \\
\midrule
CNN       & False         & $+1.27\%$ & $-1.56\%$  & $+5.27\%$ \\
CNN       & True         & $+6.40\%$ & $+1.36\%$ & $+17.35\%$ \\
LSTM    & False          & $+1.26\%$ & $-1.54\%$  & $+5.15\%$ \\
LSTM    & True          & $+6.69\%$  & $+1.47\%$ & $+17.99\%$  \\
LiBERT    & False        & $+1.90\%$ & $-1.14\%$  & $-6.16\%$\\
LiBERT    & True          & $+8.35\%$  & $+2.46\%$ & $+20.60\%$  \\
\bottomrule
\end{tabular}
\vspace{-1mm}
\end{table}

In addition, we'd like to analyze the effect of traditional features on the deep models. In As shown in Table~\ref{table:qim-trad-feat}, without traditional features, the deep models (CNN, LSTM, and LiBERT) bring much less performance gain compared with those that incorporate these features in a wide-and-deep fashion. 


\subsubsection{Online Experiments}
\label{section:scalability}

\begin{table}
\small
\caption{Online performance comparison for word-level complete query intent model vs production baseline in SERP blending.}
\label{table:qim-cnn-online}
\begin{tabular}{llc}
\toprule
\textbf{Model} & \textbf{Metrics} & \textbf{\% lift} \\
\midrule
\multirow{3}{*}{CNN-word} & CTR@5 & neutral \\
& CTR@5 for job results & +0.43\% \\
 & SAT click & neutral \\
\midrule
\multirow{3}{*}{LiBERT (vs. CNN-word)} & CTR@5 & +0.17\% \\
 & CTR@5 for job results & +0.51\% \\
 & SAT click & +1.36\% \\
\bottomrule
\end{tabular}
\end{table}

It is worth noting that even though the LSTM model performs slightly better than the CNN model, we proceed to implement the CNN model in online production. This is due to the fact that LSTM does not bring much relevance gain while introducing almost 2 times inference latency (Table \ref{table:latency}).

Table \ref{table:qim-cnn-online} shows the online metrics gain of CNN model over logistic regression. It improves two job search related metrics: the overall click-through rate of job postings at top 1 to 5 positions in the search result page (CTR@5, which is the CTR at position 1 to 5 for Job Results), and total number of job search result viewers (Job  Viewers). This is consistent with the significantly improved F1 score of job intent (offline results in Table \ref{table:qim-offline}).

Later, we conducted online experiments of LiBERT over CNN (Table \ref{table:qim-cnn-online}). Since LiBERT can further improve the accuracy of all intents, more online metrics are significant, including SAT clicks (the satisfactory clicks on documents with a long dwell time), and CTR@5 over all documents.



\subsection{Scalability}



In the online system, latency at 99 percentile (P99) is the major challenge for productionization. For query intent, offline pre-computation of the frequent queries does not help, since the P99 queries are usually rare, and there are large number of unseen queries each day at LinkedIn search. 
The latency comparison of the character-level models (CNN-char and LSTM-char) for incomplete queries is shown in Table \ref{table:latency}. The P99 latency numbers are computed on Intel(R) Xeon(R) 8-core CPU E5-2620 v4 @ 2.10GHz machine with 64-GB memory.

In terms of complete queries, the CNN/LSTM models have even less latency, since number of words are generally smaller than number of characters. For BERT models, the latency increases significantly. It is worth noting that by pre-training a smaller BERT model on in-domain data, we are able to reduce the latency from 53ms (BERT-Base) to 15ms (LiBERT) without hurting the relevance performance, which meets the search blending latency requirement.



\begin{table}
\small
\vspace{-1mm}
\caption{Latency and model size comparison on different query intent models at 99 percentile.}
\label{table:latency}
\begin{tabular}{llcc}
\toprule
  & \textbf{Model} & \textbf{\#Params} & \textbf{P99 Latency}  \\
\midrule
\multirow{3}{*}{\begin{tabular}[c]{@{}c@{}}Incomplete\\ Query Intent\end{tabular}} & Tri-letter & - &  - \\
& CNN-en           &    123k    &  $+0.38ms$  \\
& LSTM-en          &   379k   & $+2.49ms$  \\
& LSTM-i18n-embed     &   1M   & $+2.72ms$ \\
\midrule
\multirow{4}{*}{\begin{tabular}[c]{@{}c@{}}Complete\\ Query Intent\end{tabular}} & LR-BOW  & - &  - \\
& CNN-word         &   6.5M      & $+0.45ms$   \\
& LSTM-word        &   6.5M      & $+0.96ms$  \\
& BERT-Base~\cite{devlin2018bert}        &   110M     & $+53.2ms$  \\
& LiBERT          &   10M   & $+15.0ms$  \\

\bottomrule
\end{tabular}
\vspace{-3mm}
\end{table}


%% file: lesson.tex

\section{Lessons Learned}

We would like to share our experiences and lessons learned through our journey of leveraging state-of-the-art technologies and deploying the deep models for query intent prediction to online production systems. We hope they could benefit others from the industry who are facing similar tasks.

\vspace{1mm}
\noindent\textbf{Online Effectiveness/Efficiency Trade-off.} 
Our experiments with BERT-based models show that pre-training with \textit{in-domain} data improved fine-tuning performances compared to  the off-the-shelf models significantly. Aside from relevance gains that BERT brings us, latency is a big issue for productionizing these large models. We found that \textit{reducing the pre-trained model sizes} significantly reduces the inference latency while providing accurate predictions.

For the incomplete query intent, we cannot deploy complicated models due to the latency constraint in typeahead search. In order to maximize the relevance performance within a strict latency constraint, we investigated the characteristics of LSTM and CNN models w.r.t. effectiveness and efficiency. LSTMs give \textit{superior relevance performance} for its ability to capture long range information in sequences, especially when the sequence is incomplete, whereas CNNs are \textit{much faster} in inference speed by design. We were able to deploy compact LSTM models to the typeahead product while achieving an optimal prediction performance with inference latency being tolerable in production.

\vspace{1mm}
\noindent\textbf{Token Granularity.}
In the incomplete query intent models for the typeahead search product, the choice of character-level models is effective in modeling the character sequences in the incomplete queries. It's worth mentioning that these models require a smaller vocabulary size compared to word or subword level models, which result in much more \textit{compact models} suitable for online serving. 

The character-level granularity allows for combining vocabularies from different languages since many languages share similar characters. The design of multilingual models could further benefit online systems for \textit{robust predictions} across multiple markets yet \textit{easier maintenance} than per-language models.


\vspace{1mm}
\noindent\textbf{Combining Traditional Features.}
Traditional features are informative for promoting deep query intent understanding. From our experiments we show that simply discarding the handcrafted features and user personalized features and replacing them with deep learning models hurts the relevance performances. We find that incorporating the traditional features in a wide-end-deep fashion is crucial in successful intent prediction. 

%% file: related-work.tex
\section{Related Work}


Query intent prediction has been an important topic in modern search engines \cite{kang2003query,li2008learning,hu2009understanding,santos2011intent}. We firstly introduce traditional methods, then show how deep learning models are applied to this problem. Finally, we discuss the existing works regarding incomplete query intent modeling.

\vspace{-.1mm}
\subsection{Traditional Methods}
Early query intent works use rule based methods \cite{jansen2008determining,rose2004understanding}, as a high precision low recall strategy. 
However, it is hard to maintain as the rules become more complicated, and the recall is low.

More recently, statistical models have shown significant improvements in unsupervised (query clustering) \cite{ren2014heterogeneous, cheung2012sequence, aiello2011behavior} or supervised (query classification) approaches. In supervised methods, common features include unigram, language model scores, lexicon matching features, etc \cite{arguello2009sources, ashkan2009classifying, cao2009context}.

\vspace{-1mm}
\subsection{Deep Learning for Query Intent Classification}

Deep learning approaches have shown significant improvement in text classification tasks, where multiple CNN and LSTM based methods \cite{kim2014convolutional, kalchbrenner2014convolutional, lee-dernoncourt-2016-sequential,zhou2015c,liu2016} have been proposed. The closest related work is an CNN-based approach \cite{hashemi2016query} for classifying types of web search queries, with similar network architecture as in \cite{kim2014convolutional}.
It is worth noting that although the most recent technology on \cite{devlin2018bert} has shown promising improvement for text understanding including intent classification \cite{chen2019bert}, the performance on short text such as search queries is not clear. 

To our best knowledge, we have yet to see BERT models applied to real-world search systems for query intent prediction.


\subsection{Incomplete Query Intent}
There have been character-level models for classic NLP tasks such as text classification and language modeling \cite{joulin2016bag,zhang2015character,kim2016character}. 
However, we have yet to find previous work of character-level models on \textbf{incomplete} query intent prediction in search productions. We further extended our approach to a multilingual model where one model could serve many languages in international markets.

One related topic to typeahead search result blending is query auto completion \cite{Cai2016}. However, the two topics are fundamentally different in terms of system architecture. The former has complicated indexing systems to retrieve documents. A ranking system is build to blend the documents where query intent is a feature. The latter's objective is only to generate a complete query without any document side information.